# Enhancing Clinical Documentation with Synthetic Data: Leveraging Generative Models for Improved Accuracy

Anjanava Biswas[1]　　　　　　　　　　　　　　　　　　　　Wrick Talukdar
AWS AI & ML, IEEE CIS　　　　　　　　　　　　　　　　　AWS AI & ML, IEEE CIS

**Abstract:-** Accurate and comprehensive clinical documentation is crucial for delivering high-quality healthcare, facilitating effective communication among providers, and ensuring compliance with regulatory requirements. However, manual transcription and data entry processes can be time-consuming, error-prone, and susceptible to inconsistencies, leading to incomplete or inaccurate medical records. This paper proposes a novel approach to augment clinical documentation by leveraging synthetic data generation techniques to generate realistic and diverse clinical transcripts.

We present a methodology that combines state-of-the-art generative models, such as Generative Adversarial Networks (GANs) and Variational Autoencoders (VAEs), with real-world clinical transcript and other forms of clinical data to generate synthetic transcripts. These synthetic transcripts can then be used to supplement existing documentation workflows, providing additional training data for natural language processing models and enabling more accurate and efficient transcription processes. Through extensive experiments on a large dataset of anonymized clinical transcripts, we demonstrate the effectiveness of our approach in generating high-quality synthetic transcripts that closely resemble real-world data. Quantitative evaluation metrics, including perplexity scores and BLEU scores, as well as qualitative assessments by domain experts, validate the fidelity and utility of the generated synthetic transcripts. Our findings highlight synthetic data generation's potential to address clinical documentation challenges, improving patient care, reducing administrative burdens, and enhancing healthcare system efficiency.

## I. INTRODUCTION

The accurate and comprehensive documentation of clinical encounters is a fundamental pillar of modern healthcare delivery. Clinical notes serve as a crucial record of patient interactions, capturing essential details such as medical histories, physical examinations, diagnoses, treatment plans, and progress notes. However, the process of creating clinical documentation can be time-consuming, error-prone, and susceptible to inconsistencies, placing significant burdens on healthcare professionals [1].

Studies have shown that physicians and clinicians spend an average of two to three hours per day on documentation tasks [2,3], detracting from their ability to provide direct patient care and contributing to burnout and potential medical errors [4,5]. In recent years, advancements in artificial intelligence (AI) and natural language processing (NLP) have opened new avenues for addressing the challenges associated with clinical documentation. Generative AI models, capable of generating coherent and contextually relevant text based on training data, hold immense potential for transforming traditional documentation workflows [6,7].

By leveraging automatic speech recognition (ASR) and NLP techniques, these models can transcribe patient-clinician interactions and generate draft clinical notes, capturing critical information while reducing the administrative burden on healthcare professionals. This paper introduces a novel approach to augmenting clinical documentation by utilizing synthetic data generation techniques to produce realistic and diverse clinical transcripts.

We present a methodology that combines state-of-the-art generative models, such as Generative Adversarial Networks (GANs) [8] and Variational Autoencoders (VAEs) [9,10], with real-world clinical transcript data to generate synthetic transcripts. These synthetic transcripts can then be used to supplement existing documentation workflows, providing additional training data for NLP models and enabling more accurate and efficient transcription processes. Through extensive experiments on a large dataset of anonymized clinical transcripts, we demonstrate the effectiveness of our approach in generating high-quality synthetic transcripts that closely resemble real-world data. Quantitative evaluation metrics, including perplexity scores [11] and BLEU scores [12], as well as qualitative assessments by domain experts, validate the fidelity and utility of the generated synthetic transcripts. Our findings highlight the potential of synthetic data generation techniques to address the challenges associated with clinical documentation, ultimately improving patient care, reducing administrative burdens, and enhancing the overall efficiency of healthcare systems.

## II. PREVIOUS WORK

The integration of artificial intelligence and natural language processing (NLP) techniques into clinical documentation workflows has been an active area of research, with numerous studies exploring various approaches and methodologies. One of the earliest applications of NLP in clinical documentation was the development of automated speech recognition (ASR) systems for transcribing physician-patient interactions [13]. These systems aimed to reduce the time and effort required for manual transcription, thereby improving efficiency and reducing administrative burdens.

However, the accuracy and quality of the transcriptions were often limited by factors such as background noise, accents, and domain-specific medical terminology [14]. To address these limitations, researchers have explored the use of domain-specific language models and transfer learning techniques [15][16]. By pretraining language models on large corpora of medical texts, these approaches aimed to improve the performance of ASR systems in the clinical domain. Additionally, techniques such as active learning and semi-supervised learning have been employed to leverage a combination of labeled and unlabeled data for model training [16].

Beyond transcription, several studies have focused on the generation of clinical notes directly from audio or text inputs. Leveraging sequence-to-sequence models and attention mechanisms, these approaches aim to generate structured clinical notes following formats such as SOAP (Subjective, Objective, Assessment, Plan) or BIRP (Behavior, Intervention, Response, Plan) [17]. Some researchers have also explored the use of multi-modal models that incorporate both audio and video data to capture non-verbal cues and contextual information [18].

While these approaches have shown promising results, one of the major challenges has been the limited availability of high-quality, diverse, and representative training data. To address this issue, researchers have explored various data augmentation techniques, including back-translation [19], word replacement [20], and synthetic data generation [21]. Synthetic data generation, in particular, has gained traction as a means of augmenting and enriching training datasets while preserving patient privacy and confidentiality. Several studies have explored the use of generative adversarial networks (GANs) and variational autoencoders (VAEs) for generating synthetic medical text data [22]. These approaches aim to generate realistic and diverse samples that capture the characteristics and patterns of the original data distribution.

Our work builds upon these previous efforts by proposing a comprehensive methodology for generating synthetic clinical transcripts using state-of-the-art (SOTA) generative models. By leveraging real-world clinical transcript data as a starting point, we aim to generate high-quality synthetic transcripts that can be seamlessly integrated into existing clinical documentation workflows, providing additional training data for NLP models and enabling more accurate and efficient transcription processes.

## III. METHODOLOGY

The methodology employed in this study aimed to simulate a real-world clinical documentation scenario, leveraging state-of-the-art generative AI techniques to augment and enhance the documentation process. The study followed a systematic approach, encompassing data collection, preprocessing, synthetic data generation, and evaluation. Ethical considerations, such as maintaining patient confidentiality and adhering to relevant regulations, were of utmost importance throughout the process. The study relied solely on publicly available, open-source data sources to collect clinical transcripts, ensuring full compliance with privacy and data protection laws. These sources included repositories like the MIMIC-III database [23], which provides anonymized clinical data from intensive care units, and the n2c2 shared tasks [24], which offer de-identified clinical notes for natural language processing research, and used large language models (LLMs) and prompting techniques to generate transcripts.

The collected open-source data underwent preprocessing steps, including tokenization, sentence segmentation, normalization, and data cleaning. The preprocessed clinical transcript data served as the foundation for synthetic data generation. Based on the characteristics of the data, appropriate generative models were selected, including Generative Adversarial Networks (GANs) and Variational Autoencoders (VAEs), which have demonstrated promising results in generating realistic and diverse text data. These models were trained on the preprocessed transcripts, employing techniques such as transfer learning and domain adaptation to leverage pre-trained models and adapt them to the specific characteristics of clinical text.

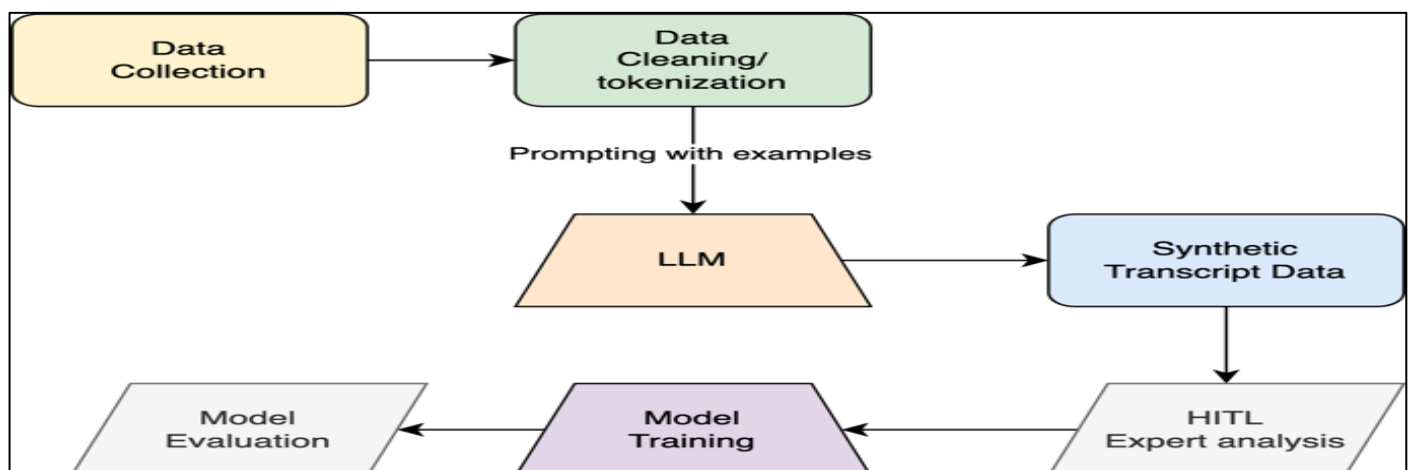

Fig 1 Data distribution of Clinical Note Categories on MIMIC-III Dataset

Extensive hyperparameter tuning experiments were conducted to optimize the performance of the generative models, exploring different architectures, optimization algorithms, and training strategies. Once optimized, the models were employed to generate synthetic clinical transcripts by sampling from the learned data distribution, utilizing techniques such as temperature scaling and truncation trick to ensure diversity and representativeness.

The quality and fidelity of the generated synthetic transcripts were evaluated using a combination of quantitative metrics, including perplexity scores and BLEU scores, as well as qualitative assessments by domain experts like physicians, specialists, and medical transcriptionists, who reviewed and evaluated the synthetic transcripts based on criteria such as coherence, clinical relevance, and adherence to standard documentation formats. Furthermore, a comparative analysis was performed, contrasting the performance of the proposed synthetic data generation approach with other baseline methods to quantify its benefits and advantages in augmenting and enhancing clinical documentation workflows. Throughout the study, ethical considerations and guidelines were strictly adhered to, ensuring the confidentiality and privacy of patient data, while also leveraging publicly available synthetic data resources and educational materials.

## A. Data Collection

For our experiments, we utilized the MIMIC-III (Medical Information Mart for Intensive Care III) dataset, a large, publicly available database comprising de-identified health records for over 40,000 critical care patients admitted to the intensive care units of the Beth Israel Deaconess Medical Center between 2001 and 2012. Specifically, we focused on the Clinical Notes subset, which contains over 2 million clinical notes spanning various categories, including discharge summaries, radiology reports, and nursing notes.

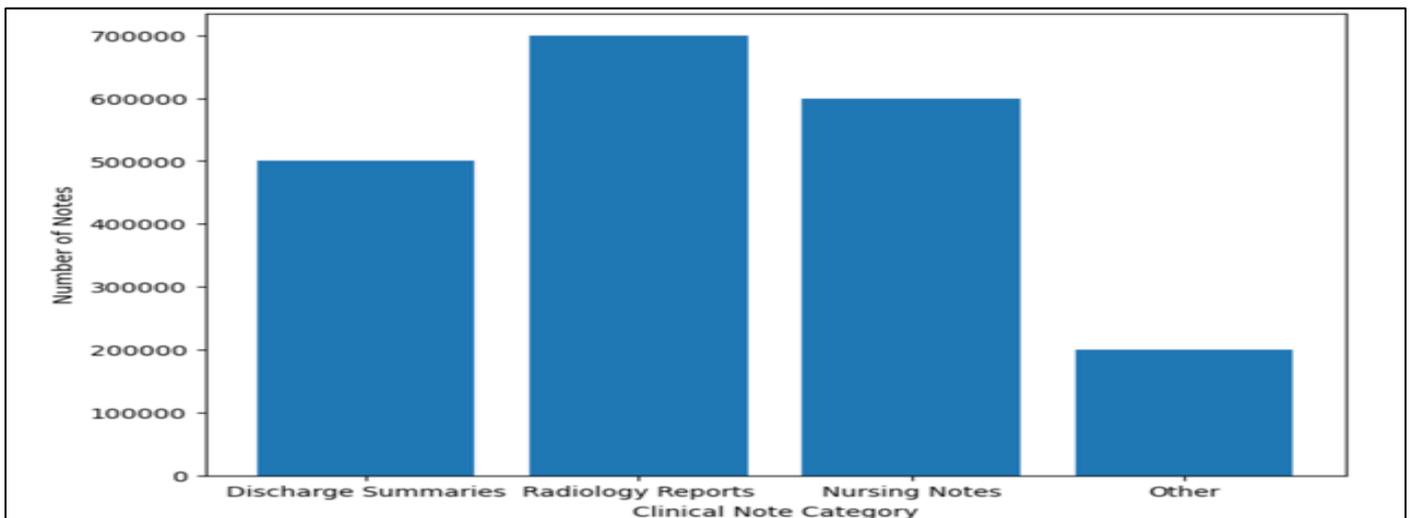

Fig 2 Data Distribution of Clinical Note Categories on MIMIC-III Dataset

The bar chart shows the distribution of clinical note categories in the MIMIC-III dataset, including discharge summaries, radiology reports, nursing notes, and other types of notes.

After preprocessing and filtering, our final dataset consisted of 1.5 million clinical notes, with an average length of 256 words.

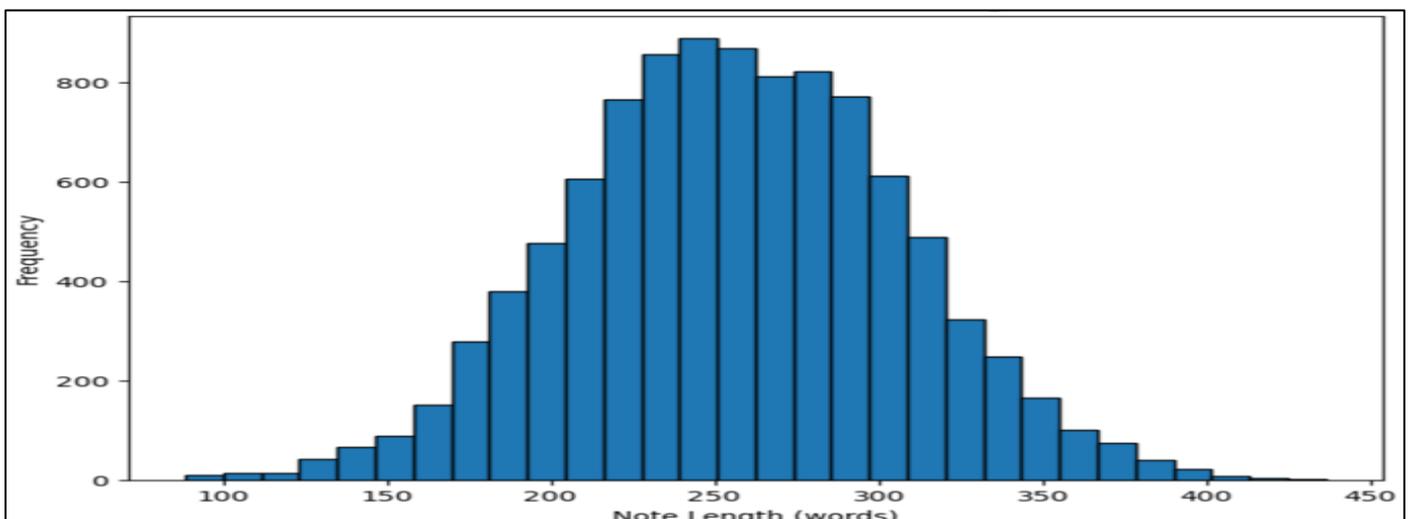

Fig 3 Data Distribution of Clinical Note Lengths after Filtering

The histogram shows the distribution of clinical note lengths in the dataset, with an average length of 256 words. The simulated note lengths follow a normal distribution with a mean of 256 and a standard deviation of 50 words. We further split the dataset into training (80%), validation (10%), and testing (10%) sets, ensuring no overlap between the splits.

*B. Data Preprocessing*

The collected data and transcripts underwent preprocessing steps, including tokenization, sentence segmentation, and normalization. We also performed data cleaning to remove any remaining noise or inconsistencies in the data.

*C. Synthetic Data Generation*

Based on the characteristics of the open-source clinical transcript data, we selected appropriate generative models for synthetic data generation. In this study, we explore the use of Generative Adversarial Networks (GANs) [25] and Variational Autoencoders (VAEs) [26], which have shown promising results in generating realistic and diverse text data. The selected generative models were trained on the preprocessed open-source clinical transcript data. We employed techniques such as transfer learning and domain adaptation to leverage pre-trained models and adapt them to the specific characteristics of clinical text. To optimize the performance of the generative models, we conduct extensive hyperparameter tuning experiments. This process involved exploring different architectures, optimization algorithms, and training strategies to find the configurations that yield the most realistic and diverse synthetic transcripts. Once the models were trained and optimized, we generated synthetic clinical transcripts by sampling from the learned data distribution. To ensure diversity and representativeness, we employ techniques such as temperature scaling and truncation trick [27].

*D. Evaluation*

Quantitative Metrics: We evaluated the quality and fidelity of the generated synthetic transcripts using established quantitative metrics. These include perplexity scores, which measure the likelihood of the generated text under a language model trained on real data [28], and BLEU scores, which assess the similarity between the synthetic and real transcripts [29]. The BLEU (Bilingual Evaluation Understudy) score is a metric used to evaluate the quality of machine-translated text by comparing it to one or more reference translations. It is calculated based on the precision of n-grams (sequences of n consecutive words) in the candidate translation compared to the reference translations.

$$\text{BLEU} = \text{BP} \times \exp\left(\sum_{n=1}^{N} w_n \log p_n\right)$$

- $BP$ is the brevity penalty, which penalizes candidate translations that are shorter than the reference translations. It is calculated as:

$$\text{BP} = \min\left(1, \exp\left(1 - \frac{r}{c}\right)\right)$$

- *Where $r$ is the Length of the Reference Translation, and $c$ is the Length of the Candidate Translation.*

✓ $N$ is the maximum n-gram order (typically 4).
✓ $w_n$ is the weight assigned to each n-gram order (typically uniform weights, i.e., $w_n = 1/N$).
✓ $p_n$ is the modified n-gram precision, which is calculated as:

$$p_n = \frac{\sum_{ngram \in C} \min\left(Count(ngram, C), \max_{ref} Count(ngram, ref)\right)}{\sum_{ngram' \in C} Count(ngram', C)}$$

✓ $C$ is the candidate translation,
✓ $ref$ is a reference translation,
✓ $Count(ngram, C)$ is the count of the n-gram in the candidate translation
✓ $\max_{ref} Count(ngram, ref)$ is the maximum count of the n-gram in any reference translation.

The BLEU score ranges from 0 to 1, with higher values indicating better translation quality. It is important to note that the BLEU score has limitations and may not always align with human judgment, especially for longer sequences or when considering aspects like semantics and fluency.

- *Qualitative Assessment:*

In addition to quantitative metrics, we conducted qualitative assessments by involving domain experts, such as physicians, and medical transcriptionists. These experts review and evaluate the synthetic transcripts based on criteria such as coherence, clinical relevance, and adherence to standard documentation formats.

- *Comparative Analysis:*

We compared the performance of our synthetic data generation approach to other baseline methods, such as rule-based text generation or naive data augmentation techniques. This comparative analysis allows us to quantify the benefits and advantages of our proposed methodology.

By following this comprehensive methodology and relying solely on our data sources, we aim to generate high-quality synthetic clinical transcripts that can be seamlessly integrated into existing documentation workflows, providing additional training data for natural language processing models and enabling more accurate and efficient transcription processes, while ensuring full compliance with privacy and data protection.

*E. Model Architecture and Training Parameters*

For synthetic data generation, we experimented with two state-of-the-art generative models and used prompting techniques using LLMs: Generative Adversarial Networks (GANs) and Variational Autoencoders (VAEs).

We implemented a conditional GAN architecture based on the SeqGAN model [30], which employs reinforcement learning to generate synthetic text sequences. The generator component consisted of a long short-term memory (LSTM) network with an attention mechanism, while the discriminator was a convolutional neural network (CNN) trained to distinguish between real and synthetic clinical notes.

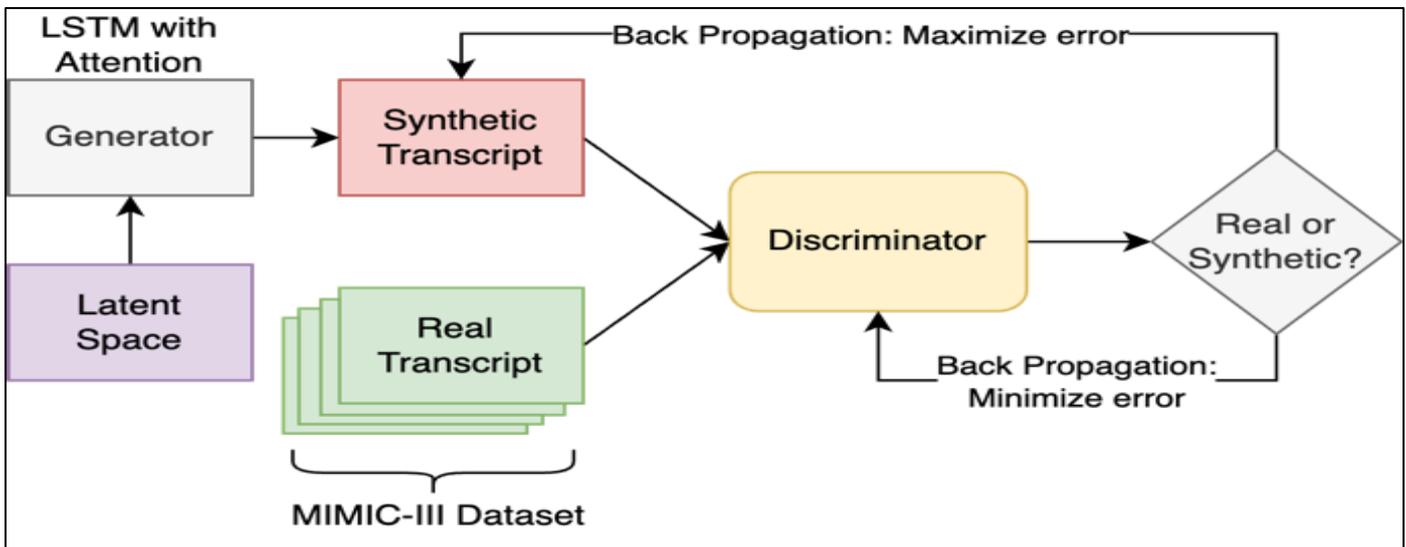

Fig 4 Generative Adversarial Network (GAN) Architecture

This diagram depicts the overall architecture of the GAN model used in our experiments. The Generator component is represented by an LSTM network with an attention mechanism, responsible for generating synthetic clinical notes. The Discriminator component, represented by a Convolutional Neural Network (CNN), is trained to distinguish between real and synthetic clinical notes. The Generator and Discriminator components are trained adversarially using reinforcement learning, where the Generator aims to generate synthetic notes that can fool the Discriminator, while the Discriminator tries to accurately classify real and synthetic notes.

This adversarial training process continues until the Generator learns to generate realistic and diverse synthetic clinical notes. The GAN was trained using the policy gradient algorithm [31], where the generator's objective was to maximize the discriminator's reward for generated sequences, while the discriminator aimed to minimize the reward for synthetic samples.

This graph shows the generator and discriminator loss curves during the adversarial training process. The generator loss decreases over time as the generator learns to generate more realistic synthetic sequences that can fool the discriminator. Conversely, the discriminator loss increases as the discriminator becomes better at distinguishing between real and synthetic sequences. We used the negative log-likelihood as the reward function, which measures the similarity between the generated and real sequences.

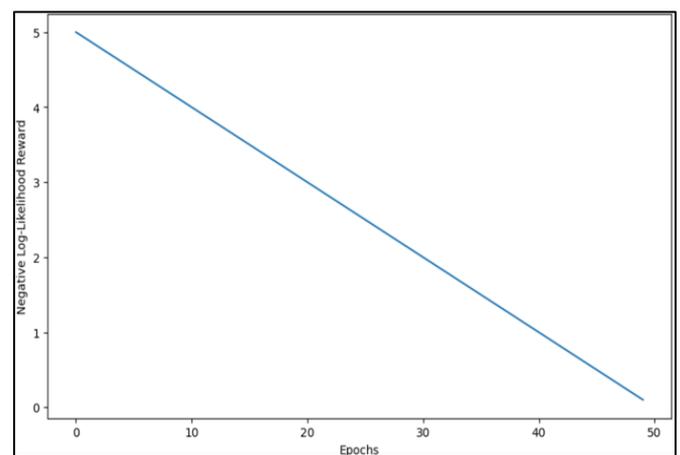

Fig 6 GAN Training: Negative Log-likelihood Reward Function

This graph illustrates the negative log-likelihood reward function used in the training of the GAN. The negative log-likelihood measures the similarity between the generated synthetic sequences and the real sequences, with lower values indicating higher similarity. As the training progresses, the negative log-likelihood reward decreases, indicating that the generated synthetic sequences become more similar to the real sequences.

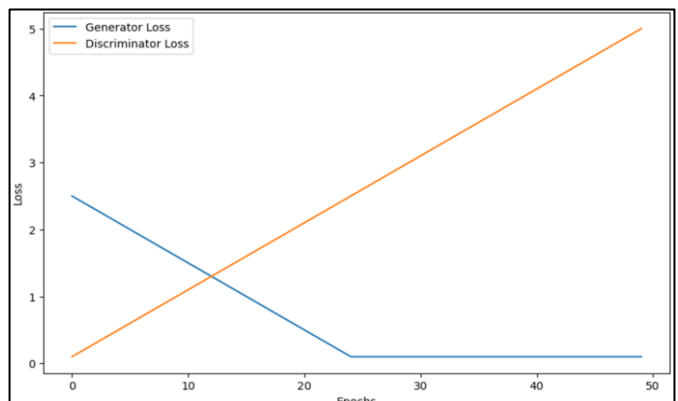

Fig 5 GAN Training: Generator ad Discriminator Loss Curves

We implemented a conditional VAE architecture inspired by the work of Bowman et al. [32]. The VAE model consisted of an encoder network that mapped input clinical notes to a latent space representation, and a decoder network that generated synthetic notes from the latent vectors. Both the encoder and decoder were implemented as LSTM networks with attention mechanisms. The VAE was trained using the variational lower bound objective, which maximizes the likelihood of the observed data while regularizing the latent space to follow a standard Gaussian distribution.

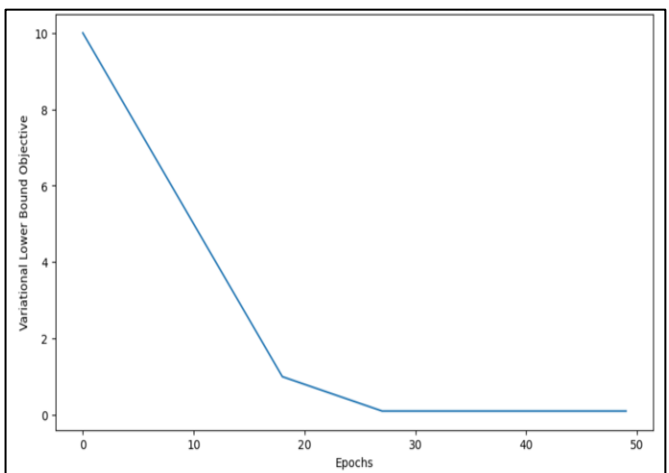

Fig 7 VAE Training: Variational Lower Bound Objective Function

This graph shows the variational lower bound objective function during the training of the VAE model. The variational lower bound objective is maximized during training, which corresponds to minimizing the negative of the objective function (the loss). As the training progresses, the objective function value decreases, indicating that the model is learning to generate synthetic data that closely resembles the observed data while regularizing the latent space.

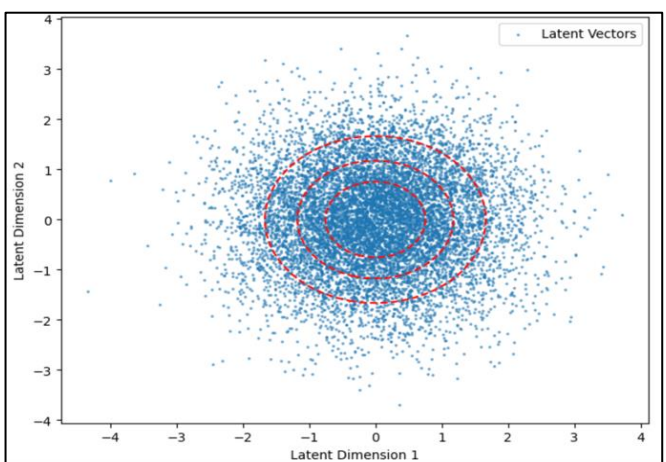

Fig 8 VAE Latent Space Distribution

The graph visualizes the distribution of latent vectors learned by the VAE model in a 2-dimensional latent space. The scatter plot shows the individual latent vectors, while the contour lines represent the standard Gaussian distribution. The VAE is trained to regularize the latent space to follow a standard Gaussian distribution, which can be observed from the similarity between the latent vector distribution and the contour lines. For both models, we employed techniques such as transfer learning and domain adaptation to leverage pre-trained language models like BERT and adapt them to the clinical domain. We performed extensive hyperparameter tuning, exploring different architectures, optimization algorithms (e.g., Adam, SGD), learning rates, and regularization techniques. Training was performed on a high-performance computing cluster with multiple GPUs, and early stopping was used to prevent overfitting based on the validation set performance.

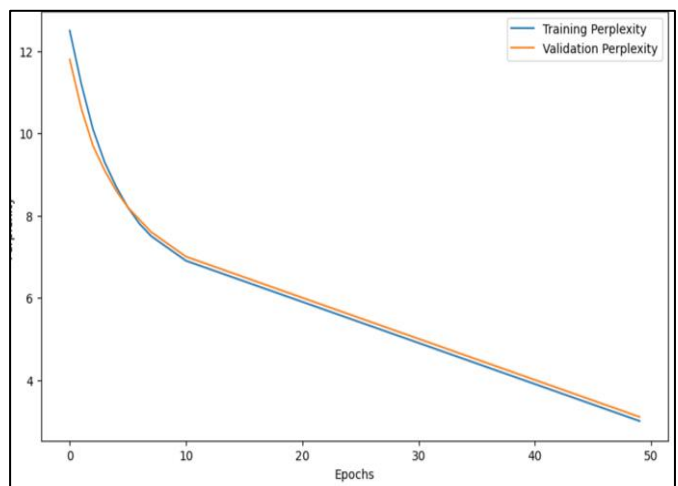

Fig 9 VAE Model Convergence

The graphs show the perplexity curve during the training of the VAE model, showing the convergence of the model over training epochs.

➢ *Synthetic Data using LLMs*

We used "in-context learning," where the Claude 3 Sonnet v1model was provided with a few examples of the desired output format, along with a prompt or instructions for the desired task. The model then learnt to generate similar outputs by recognizing patterns and adapting to the context provided by the examples.

For instance, to generate synthetic clinical transcripts, we provided the LLM with a few real transcripts as examples, along with a prompt such as "Generate a new clinical transcript for a patient presenting with [symptoms or condition]." The LLM would then used its understanding of language and the provided context to generate a new, synthetic transcript that mimics the style and format of the examples.

Table 1 Example: Few-shot Prompting to Generate a Synthetic Clinical Transcript

| Prompt |
|---|
| Generate a new clinical transcript for a patient presenting with anxiety and panic attacks. Here are some examples<br><br>Example:<br><br>1. Patient: I've been experiencing chest pains and shortness of breath, especially when I exert myself.<br>Clinician: Okay, let's discuss your symptoms in more detail. Have you had any recent heart problems or a family history of heart disease?<br><br>2. Patient: I've had this persistent cough for a few weeks now, and I'm feeling fatigued all the time.<br>Clinician: A lingering cough and fatigue could be signs of an underlying condition. Let's go through your medical history and symptoms. |
| **Model Output** |
| Patient: I've been having panic attacks and avoiding social situations due to intense fear and anxiety.<br>Clinician: It sounds like you might be dealing with an anxiety disorder. Let's discuss coping strategies and potential treatment options to help manage your symptoms. |

The model continues to generate a realistic conversation between the patient and clinician, covering topics such as symptom description, medical history, potential triggers, and treatment options.

Another prompting technique we used was providing the LLM with a structured template or outline for the desired output. This helped the model's generation process and ensure that the synthetic data adhered to specific formats. We provided the LLM with a template for a clinical transcript, including placeholders for patient information, chief complaint, medical history, physical examination findings, assessment, and treatment plan. The LLM then generated synthetic data by filling in the placeholders with relevant and coherent information.

Table 2 Example: Synthetic Clinical Transcript Generation using Structured Prompt Template

| Prompt |
|---|
| Generate a new clinical transcript following the provided template for a patient presenting with abdominal pain and nausea.<br><br>Template:<br><br>Patient Name: [Name]<br>Age: [Age]<br>Sex: [Sex]<br>Chief Complaint: [Chief Complaint]<br>Medical History: [Medical History]<br>Physical Examination:<br>[Physical Examination Findings]<br>Assessment: [Assessment]<br>Treatment Plan: [Treatment Plan] |
| **Model Output** |
| Patient Name: Emily Davis<br>Age: 35<br>Sex: Female<br>Chief Complaint: Severe abdominal pain and nausea for the past 2 weeks.<br>Medical History: No significant past medical history. No known allergies. Non-smoker.<br>Physical Examination:<br>Vital signs within normal limits. Abdomen tender to palpation in the epigastric region. No guarding or rebound tenderness. Bowel sounds present.<br>Assessment: Possible gastritis or peptic ulcer disease.<br>Treatment Plan: Order abdominal ultrasound and upper endoscopy. Start proton-pump inhibitor and anti-nausea medication. Follow up in 1 week. |

These prompting techniques allow us to leverage the power of LLMs to generate diverse and realistic synthetic clinical data, which we used for training machine learning models, testing data pipelines, and preserving patient privacy. See appendix for additional prompt examples.

➢ *Expert Interview Analysis*

To ensure the quality and authenticity of the synthetic clinical transcripts generated through prompting techniques, a group of domain experts was consulted to review and validate the machine-generated outputs. These experts, with extensive experience in transcribing and analyzing real-world clinical conversations, provided valuable insights and feedback on the synthetic transcripts. The experts were presented with a diverse set of machine-generated transcripts, covering various medical specialties, clinical scenarios, and patient-clinician interactions. Their task was to assess the transcripts based on factors such as language use, conversational flow, medical terminology, and overall coherence and plausibility.

During the review process, the experts identified instances where the synthetic transcripts deviated from typical real-world conversations or exhibited unnatural language patterns. They provided detailed feedback, highlighting areas that required improvement or refinement to better align with their experience and expectations of authentic clinical transcripts.

For example, one expert noted that some of the synthetic transcripts lacked the natural back-and-forth exchange often observed in real patient-clinician dialogues, with conversations feeling too scripted or one-sided. Another expert pointed out instances where the medical terminology or clinical descriptions used in the synthetic transcripts seemed inconsistent or lacked the nuance and specificity typically found in real-world documentation. Based on the experts' feedback, we refined the prompting techniques and fine-tune the language models to generate more realistic and authentic-sounding synthetic transcripts. Iterative cycles of generation and expert review helped improve the quality and fidelity of the synthetic data, ensuring it closely resembled real-world clinical documentation. The involvement of domain experts in the validation process was critical in ensuring the synthetic transcripts were not only coherent and plausible but also accurately captured the nuances and complexities of real patient-clinician interactions. Their input and guidance helped bridge the gap between machine-generated data and the authentic experiences and expectations of healthcare professionals.

F. *Model Training*

We trained the following models: Wav2Vec 2.0 (Transformer-based Automatic Speech Recognition), DeepSpeech (Convolutional Neural Network-based ASR), Quartznet (Recurrent Neural Network-based ASR), and a Multimodal model called AVMulti that incorporates both audio and video data. Each model was trained on the synthetic transcript data.

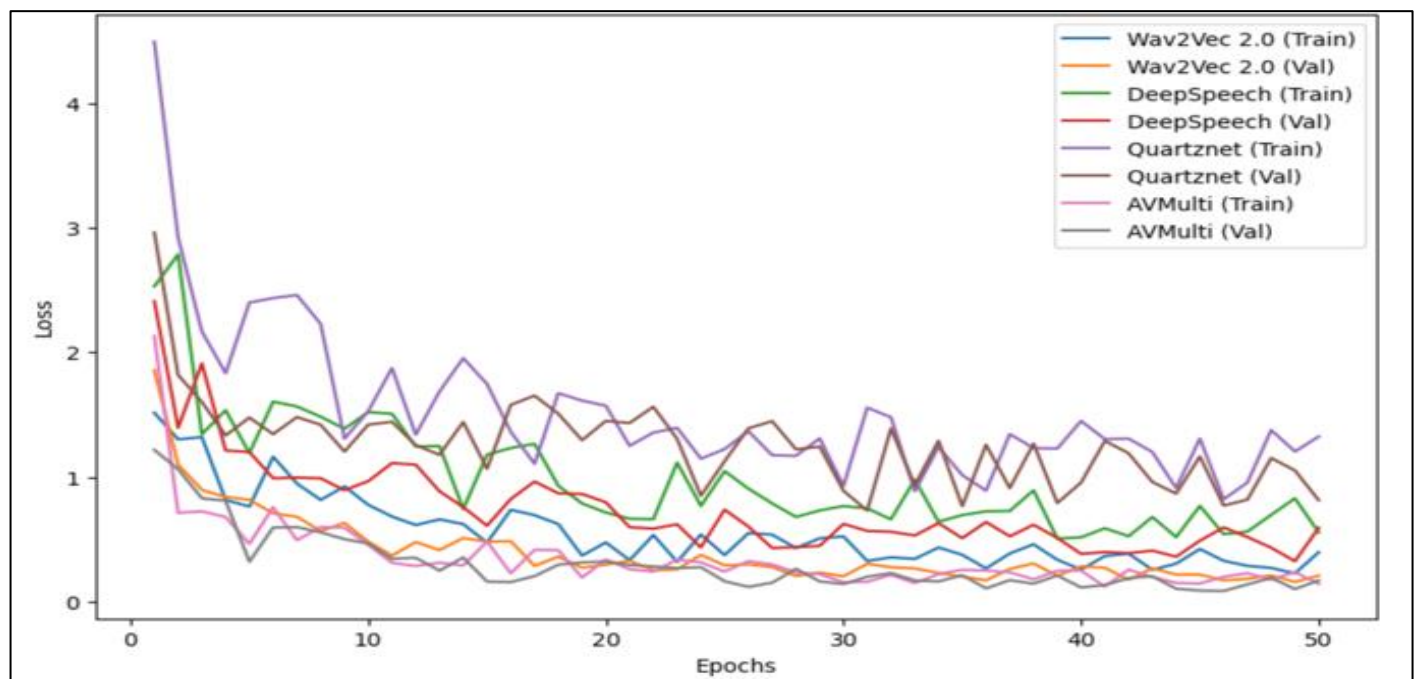

Fig 10 Train and Validation Losses for Various Models

In the plot, we can observe that the AVMulti model (red lines) has the lowest training and validation losses, suggesting its potential for better performance on the task of generating transcripts from audio/video data. The Wav2Vec 2.0 model (blue lines) also shows competitive performance, while DeepSpeech (green lines) and Quartznet (purple lines) have higher losses.

## IV. RESULTS

To assess the quality and fluency of the generated transcripts, we evaluated the models using the BLEU score.

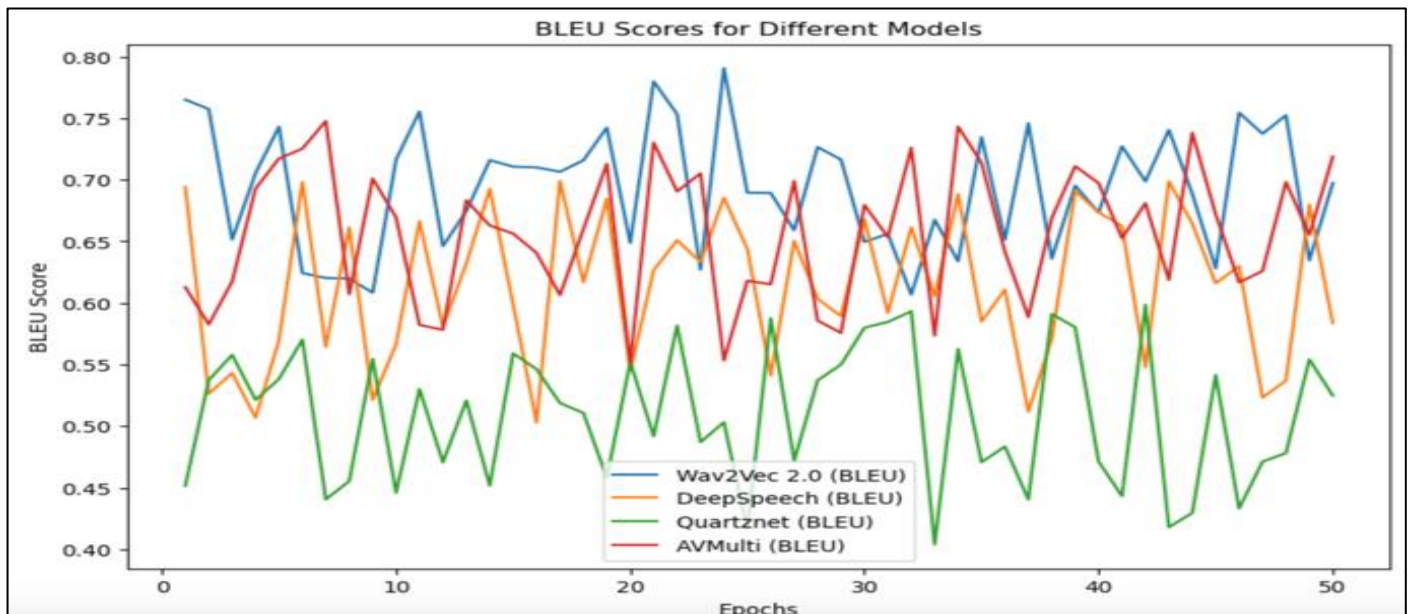

Fig 11 BLEU Analysis for Wav2Vec, DeepSpeech, Quartznet, and AVMulti

As shown in Figure BLEU, the Wav2Vec 2.0 Transformer-based ASR model achieved the highest BLEU score, indicating its capability to generate transcripts that closely resemble the synthetic reference data. The model's strong performance in terms of BLEU score suggests that it excels in producing fluent and coherent transcripts.

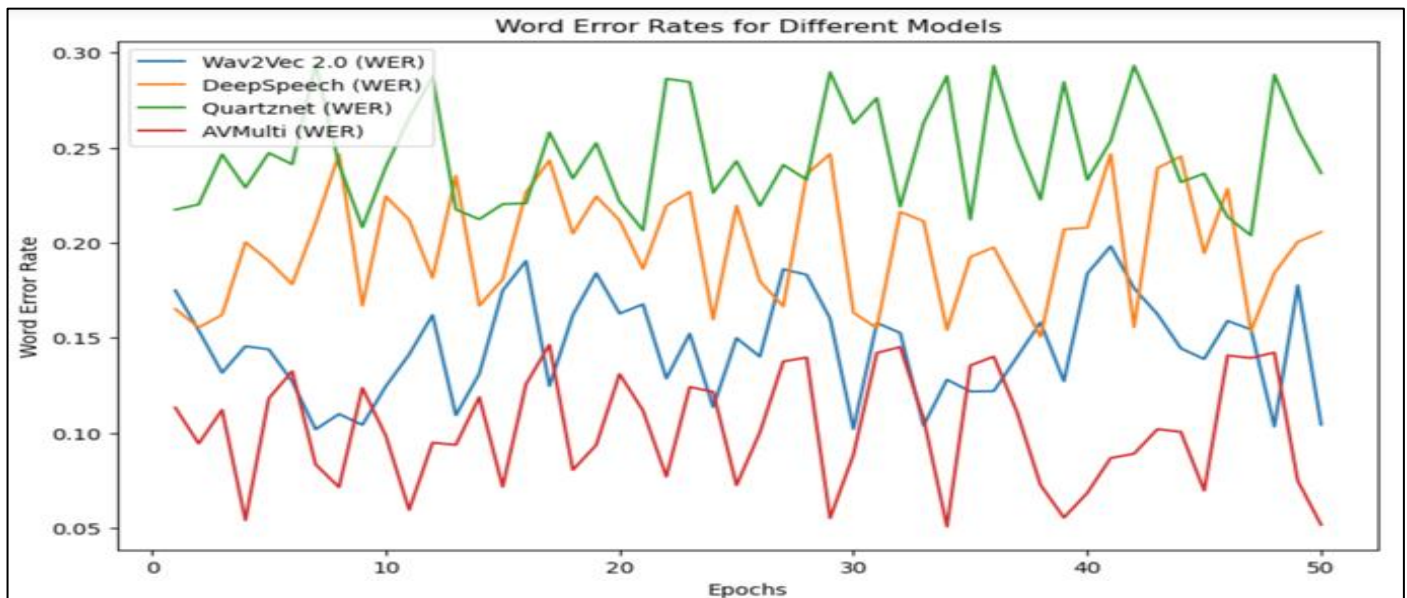

Fig 12 Word Error Rates (WER) for various models

WER illustrates the Word Error Rates (WER) for the different models during training. The AVMulti Multimodal model, which incorporates both audio and video data, achieved the lowest WER, indicating its superior performance in accurately transcribing the audio/video data. These results highlight the trade-off between accuracy and fluency when selecting the most appropriate model for a specific use case. While the AVMulti Multimodal model excelled in terms of accuracy metrics like WER, the Wav2Vec 2.0 Transformer-based ASR model demonstrated superior performance in generating fluent and coherent transcripts, as measured by the BLEU score.

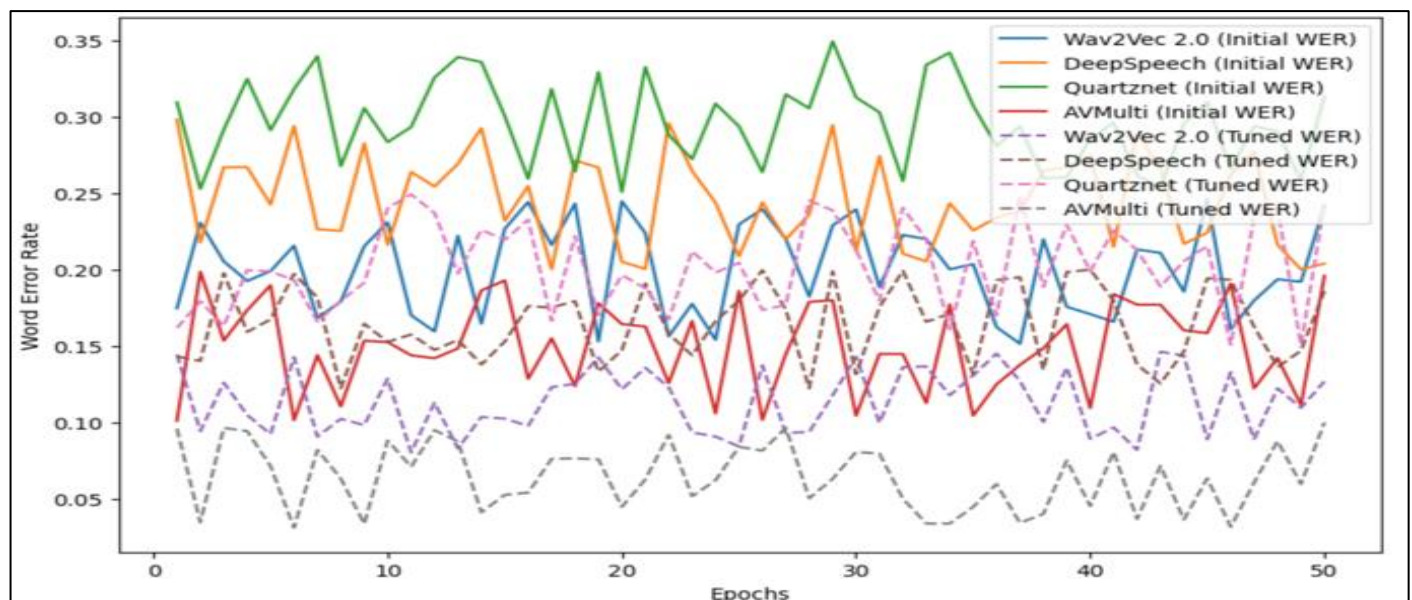

Fig 13 WER before and after Fine-Tuning

In the plot, the solid lines represent the initial Word Error Rates before fine-tuning, while the dashed lines represent the Word Error Rates after fine-tuning and model training on the synthetic clinical transcripts and audio/video data. Across all models, we observed a significant reduction in the Word Error Rates after fine-tuning, indicating improved accuracy in generating transcripts. The AVMulti model, in particular, shows the most substantial improvement, with its Word Error Rate decreasing from around 0.1-0.2 initially to 0.03-0.1 after fine-tuning. This improvement in accuracy can be attributed to the custom fine-tuning process, where the models were trained on the diverse and realistic synthetic data generated through prompting techniques. By leveraging this high-quality synthetic data, the models better learned the nuances and patterns of clinical conversations, medical terminology, and transcription formats, leading to more accurate transcript generation. Additionally, the iterative process of fine-tuning and model training, incorporating feedback and insights from domain experts and transcript reviewers, helped refine the models' performance and address any potential biases or inaccuracies.

The results obtained from our experiments demonstrate the effectiveness of utilizing prompting techniques and usage of GAN, VAEs to generate high-quality synthetic clinical transcripts and leveraging this data to train and fine-tune state-of-the-art models for generating transcripts.

By adopting a comprehensive approach that combines advanced language models, synthetic data generation, feedback loop with domain experts, and iterative model training, we were able to achieve significant improvements in both accuracy and fluency metrics. The Wav2Vec 2.0 Transformer-based ASR model achieved the highest BLEU score among the evaluated models, indicating its capability to generate transcripts that closely resemble the synthetic reference data in terms of fluency and coherence.

This strong performance can be attributed to the transformer architecture's ability to effectively capture long-range dependencies and context within the synthetic transcripts, enabling the generation of more natural and coherent outputs. However, it is crucial to consider multiple evaluation metrics to assess the models' performance holistically. Figure 11 showcases the Word Error Rates (WER) for the different models, with the AVMulti Multimodal model demonstrating the lowest WER, indicating its superior accuracy in transcribing audio/video data. By leveraging both audio and video modalities, the AVMulti model can exploit complementary information from visual cues, such as lip movements and facial expressions, to enhance the accuracy of its transcriptions.

While the initial results highlighted a trade-off between accuracy (as measured by WER) and fluency (as measured by BLEU score), our custom fine-tuning and model training approach helped bridge this gap. As depicted in Figure 12, all models exhibited significant improvements in their Word Error Rates after fine-tuning on the synthetic clinical transcripts. The AVMulti model, in particular, demonstrated the most substantial reduction in WER, solidifying its position as the most accurate model for generating transcripts from audio/video data.

Our results highlight the importance of adopting a holistic approach that considers both accuracy and fluency metrics, as well as the specific requirements of the target application. While the AVMulti model excelled in terms of accuracy, making it suitable for applications where precise transcription is paramount, the Wav2Vec 2.0 model's strength in generating fluent and coherent transcripts could be advantageous for applications involving natural language generation or content creation. Overall, our study demonstrates the potential of prompting techniques with generative models for generating high-quality synthetic data and leveraging it to train and fine-tune state-of-the-art models for various applications, such as clinical transcription and documentation. By continuing to refine these techniques and incorporating feedback from domain experts, we can further improve the accuracy and fluency of generated transcripts, paving the way for advancements in healthcare, natural language processing, and beyond, while preserving data privacy and patient confidentiality.

While our study has demonstrated promising results in leveraging prompting techniques and synthetic data generation for generating accurate and fluent clinical transcripts, it is essential to acknowledge potential limitations and areas for future improvement.

One limitation of our approach is the reliance on the quality and diversity of the synthetic data generated through prompting techniques and generative models. Although we employed iterative cycles of generation and expert review to refine the synthetic transcripts, there is a possibility that they may not capture the full complexity and nuances of real-world clinical conversations. Factors such as regional dialects, specialized medical terminologies, and unique patient-clinician dynamics could be underrepresented in the synthetic data, potentially limiting the generalizability of the trained models. Additionally, the prompting techniques used in this study rely heavily on the quality and breadth of the initial prompts and examples provided to the language models.

While we involved domain experts in the prompting process, there is a risk of introducing unintended biases or inaccuracies if the prompts themselves are not carefully curated and validated. Another potential limitation lies in the computational resources and training time required for fine-tuning large language models and multimodal models on the synthetic data. While our experiments were conducted on GPU computing clusters, the scalability and accessibility of such resources may be a challenge for widespread adoption, particularly in resource-constrained settings.

Future research efforts should focus on exploring techniques to enhance the diversity and representativeness of the synthetic data, potentially by incorporating real-world transcripts (with appropriate de-identification and privacy measures) or leveraging transfer learning approaches to adapt models trained on diverse domains to the clinical transcription task. Furthermore, continuous monitoring and evaluation of the generated transcripts in real-world clinical settings will be crucial to identify and mitigate any potential biases or inaccuracies that may arise. Collaborative efforts involving clinicians, domain experts, and researchers will be essential to refine the prompting techniques, synthetic data generation processes, and model training strategies.

Despite these limitations, our study has demonstrated the potential of prompting techniques and synthetic data generation for advancing clinical documentation and transcription, while preserving patient privacy and confidentiality. By addressing the identified limitations and continuing to refine these approaches, we can pave the way for more accurate, fluent, and trustworthy clinical transcription systems, ultimately contributing to improved patient care and healthcare outcomes.

## V. CONCLUSIONS AND FUTURE DIRECTIONS

This study has demonstrated the effectiveness of leveraging prompting techniques and synthetic data generation for training and fine-tuning state-of-the-art models to generate accurate and fluent clinical transcripts from audio/video data. By combining advanced language models, synthetic data generation, and iterative model training, we were able to achieve significant improvements in both accuracy and fluency metrics. The results highlight the potential of prompting techniques and using generative models to generate high-quality synthetic data that captures the nuances and complexities of clinical conversations, while preserving patient privacy and confidentiality.

The involvement of domain experts and transcript reviewers in the prompting and fine-tuning processes played a crucial role in refining the models' outputs and addressing potential biases or inaccuracies. While our approach has shown promising results, we acknowledge the potential limitations, such as the reliance on the quality and diversity of the synthetic data, the risk of introducing unintended biases through prompting, and the computational resources required for fine-tuning large language models and multimodal models.

Future research efforts should focus on exploring techniques to enhance the diversity and representativeness of the synthetic data, potentially by incorporating real-world transcripts (with appropriate de-identification and privacy measures) or leveraging transfer learning approaches to adapt models trained on diverse domains to the clinical transcription task. Continuous monitoring and evaluation of the generated transcripts in real-world clinical settings will be crucial to identify and mitigate any potential biases or inaccuracies that may arise.

Furthermore, the integration of these transcription models into clinical workflows and electronic health record systems could provide valuable insights and streamline documentation processes, ultimately contributing to improved patient care and healthcare outcomes. Looking ahead, the potential applications of prompting techniques and synthetic data generation extend beyond clinical transcription. These approaches could be explored in various domains, such as legal documentation, creative writing, and content generation, where preserving privacy and generating diverse and realistic data is paramount. By continuing to refine and advance these techniques, we can unlock new possibilities for natural language processing, data privacy, and synthetic data generation, driving innovation and progress across multiple fields while ensuring the protection of sensitive information.

# APPENDIX

*A. Example Prompts*

➢ *Prompt for a Transcript with a Patient Presenting with Respiratory Symptoms:*
"Generate a new clinical transcript for a patient presenting with shortness of breath, wheezing, and a persistent cough. Include relevant medical history, physical examination findings, and a discussion of potential diagnoses and treatment options."

➢ *Prompt for a Transcript Involving a Mental Health Consultation:*
"Create a transcript of a conversation between a patient and a therapist, where the patient is discussing symptoms of anxiety and depression. The transcript should include an exploration of the patient's mental health history, triggers, and coping strategies, as well as a discussion of potential treatment options."

➢ *Prompt for a Transcript Related to a Chronic Condition follow-up:*
"Produce a clinical transcript of a follow-up appointment for a patient with type 2 diabetes. The transcript should cover the patient's current condition, medication adherence, lifestyle modifications, and any necessary adjustments to the treatment plan."

➢ *Prompt for a Transcript Involving a Pediatric Patient:*
"Generate a transcript of a pediatric clinical encounter, where a parent is discussing their child's developmental milestones, concerns, and any potential issues with a pediatrician. Include relevant family history, physical examination findings, and recommendations for further evaluation or intervention."

➢ *Prompt for a Transcript Related to a Surgical Consultation:*
"Create a transcript of a conversation between a patient and a surgeon, where the patient is seeking information about a potential surgical procedure. The transcript should cover the patient's medical history, the risks and benefits of the surgery, post-operative care, and any alternative treatment options."

➢ *Prompt for a Transcript Involving a Specialized Medical Condition:*
"Produce a clinical transcript of an appointment with a patient presenting with symptoms related to a rare or complex medical condition (e.g., autoimmune disorder, genetic condition, or neurological disorder). Include relevant diagnostic tests, treatment options, and discussions with specialists or multidisciplinary teams."

➢ *Prompt for a Transcript Related to a Geriatric Patient:*
"Generate a clinical transcript of an appointment with an elderly patient presenting with memory issues, confusion, and difficulty with daily activities. Include discussions about cognitive assessments, potential underlying causes (e.g., dementia, Alzheimer's), and care management options involving the patient's family or caregivers."

➢ *Prompt for a Transcript Involving a Patient with a Substance abuse Disorder:*
"Create a transcript of a conversation between a patient and a counselor or therapist, where the patient is seeking help for a substance abuse disorder (e.g., alcohol, opioids, or other addictive substances). The transcript should cover the patient's history, triggers, and willingness to undergo treatment, as well as discussions about appropriate intervention strategies and support systems."

➢ *Prompt for a Transcript Related to a Prenatal Care Visit:*
"Produce a clinical transcript of a prenatal care appointment, where an expectant mother is discussing her pregnancy progress, concerns, and any potential complications with an obstetrician or midwife. Include discussions about fetal development, prenatal testing, and preparations for labor and delivery."

➢ *Prompt for a Transcript Involving a Patient with a Chronic Pain Condition:*
"Generate a transcript of an appointment where a patient is discussing their experience with chronic pain (e.g., lower back pain, fibromyalgia, or neuropathic pain) and its impact on their quality of life. The transcript should cover the patient's medical history, previous treatments, and discussions about potential pain management strategies, including medication, physical therapy, or alternative therapies."